\documentclass[sigconf,10pt,nonacm]{acmart}

\renewcommand\footnotetextcopyrightpermission[1]{}

\AtBeginDocument{%
  }
\usepackage{dblfloatfix} 
\usepackage{algorithm}
\usepackage{algpseudocode}
\usepackage[T1]{fontenc}

\usepackage{pifont}
\newcommand{\cmark}{\ding{51}}%
\newcommand{\xmark}{\ding{55}}%

\begin{document}

\title[INAR-VL: Input-Aware Routing]{INAR-VL: Input-Aware Routing for Edge--Cloud Vision--Language Inference}

\author{Ahmed Šabanović}
\email{ahmed.sabanovic@tuwien.ac.at}
\affiliation{%
  \institution{TU Wien}
  \city{Vienna}
  \country{Austria}}

\author{Paul Joe Maliakel}
\email{paul.maliakel@tuwien.ac.at}
\affiliation{%
  \institution{TU Wien}
  \city{Vienna}
  \country{Austria}}  
  
\author{Ivona Brandić}
\email{ivona.brandic@tuwien.ac.at}
\affiliation{%
  \institution{TU Wien}
  \city{Vienna}
  \country{Austria}}

\begin{abstract}
Edge deployment of Vision--Language Models (VLMs) faces a tradeoff between latency and accuracy: cloud execution provides high-quality predictions but incurs communication delay and energy cost, while edge-only execution is faster but less accurate due to limited model capacity. This tradeoff is further complicated by heterogeneity in image quality and reasoning complexity, making static placement suboptimal. We present INAR-VL, a lightweight edge--cloud routing system for multimodal inference in a two-tier deployment. INAR-VL maintains complementary VLMs across edge and cloud and uses lightweight image and text complexity signals to guide routing and model selection, executing simple queries locally while offloading complex ones when beneficial. Evaluation on visual question answering shows that INAR-VL executes 36\% of requests on the edge, reduces latency by 24\%, lowers energy by 26\%, and preserves 97\% of cloud-level accuracy.
\end{abstract}



\maketitle

\section{INTRODUCTION}

VLMs are increasingly deployed in edge environments such as mobile devices, robots, and embedded cameras, where inference must operate under strict constraints in latency, energy, memory, and bandwidth \cite{DBLP:journals/csr/WangGZYG25, DBLP:journals/corr/abs-2601-14921, DBLP:conf/nips/KohFS23}. Consider a field technician using smart glasses to query equipment labels and maintenance instructions while moving through a factory floor with patchy Wi-Fi. In such scenarios, sending every request to the cloud introduces unpredictable communication delays, while relying solely on a small on-device model risks incorrect answers on visually degraded inputs or complex reasoning queries. Serving VLMs efficiently in such environments remains challenging \cite{DBLP:journals/corr/abs-2405-10739}. Unlike text-only workloads, multimodal inputs exhibit high variability in image quality (e.g., blur, lighting, compression artifacts) and text complexity, which strongly affects both model accuracy and inference cost \cite{DBLP:journals/corr/abs-2509-12492}.

Existing approaches largely ignore this heterogeneity. Systems such as MoA-Off and recent routing approaches inspired by RouteLLM-style difficulty estimation rely on static deployment choices or primarily text-based signals, which are insufficient for multimodal workloads \cite{DBLP:journals/corr/abs-2509-16995, DBLP:conf/cvpr/WangLLZMWZCZLZ24, DBLP:conf/iclr/DingM0SMRLA24}. While effective for text-only LLMs \cite{DBLP:conf/iclr/DingM0SMRLA24}, they fail to capture visual degradation effects, where, for example, increasing resolution helps fine-grained images but not blurred ones \cite{DBLP:journals/corr/abs-2509-12492}. Our measurements highlight this trade-off: edge-only inference is faster and more energy-efficient but less accurate than cloud-only execution.

To make this trade-off explicit, our baselines show that edge-only execution achieves 66.5\% accuracy at 824 ms and 7.5 J per sample, while cloud-only reaches 74.4\% accuracy at 2408 ms and 26.0 J. This gap motivates input-aware routing instead of static placement.

However, designing such a router for edge environments introduces several challenges. First, different VLM architectures exhibit complementary strengths: some models are better at perception-heavy tasks such as OCR or fine-grained visual recognition, while others perform better on language-intensive reasoning queries\cite{DBLP:journals/corr/abs-2509-23661}. Second, routing decisions must be made \emph{before} expensive inference, requiring lightweight request characterization rather than additional model evaluations. Third, routing must respect practical deployment constraints including limited edge GPU memory, device energy budgets, network variability, and strict overhead limits for pre-inference analysis.

To address these challenges, we present INAR-VL, a practical routing system for adaptive edge--cloud VLM inference. Unlike prior approaches that rely on a single model or static offloading decisions, INAR-VL maintains a heterogeneous pool of complementary vision–language models distributed across edge and cloud resources. A key novelty of our approach is the use of lightweight multimodal signals—capturing both image quality and text complexity—extracted prior to inference to guide routing decisions without invoking expensive VLM computation. These signals drive a routing policy that jointly optimizes three dimensions: model architecture, input resolution, and execution placement. This joint optimization across model, resolution, and placement has not been explicitly addressed in prior VLM routing systems.

\textbf{Contributions.} We design INAR-VL, a routing system that uses lightweight multimodal signals for pre-inference request characterization, enabling model and placement decisions without additional VLM cost. The policy is deployment-aware, jointly optimizing model choice, resolution, and edge/ cloud placement under practical constraints such as memory, energy, and network variability. We further provide an end-to-end evaluation on an 8\,GB edge GPU and cloud setup across three VQA benchmarks, demonstrating improved latency--accuracy trade-offs over static baselines.

\section{BACKGROUND \& MOTIVATION}

Designing an effective routing policy requires understanding both
the hardware constraints of edge deployments and the variability
in model capabilities across different VLM architectures. In this
section, we briefly review these two aspects and highlight why
static edge or cloud placement performs poorly for heterogeneous
multimodal workloads.

\subsection{Edge Resource Constraints}
Modern VLMs combine language backbones, visual encoders, and multimodal projection layers, resulting in substantial memory and compute requirements. For example, a 7B model executed in FP16 typically requires roughly 14–16\,GB of GPU memory for weights alone, excluding activations and decoding caches \cite{DBLP:journals/csr/WangGZYG25, DBLP:journals/corr/abs-2601-14921}. These requirements exceed the capacity of many edge platforms operating under strict power and thermal limits. As a result, executing large VLMs entirely on edge hardware is often impractical, while sending every request to the cloud introduces additional network latency and bandwidth overhead. Hybrid edge--cloud deployments therefore represent a practical compromise.

\subsection{Heterogeneity Across Vision--Language Models}
VLM architectures show complementary strengths. Qwen-VL achieves \textbf{79.5\% VQAv2, 59.3\% GQA, 63.8\% TextVQA}, excelling at OCR and visually grounded language tasks \cite{DBLP:journals/corr/abs-2308-12966}, whereas LLaVA-1.5 scores \textbf{78.5\% VQAv2, 62.0\% GQA, 58.2\% TextVQA}, favoring reasoning-oriented VQA \cite{DBLP:journals/corr/abs-2410-16236}. Other studies confirm that LLaVA-OneVision-1.5 is stronger on reasoning benchmarks, while Qwen-VL outperforms on OCR tasks \cite{DBLP:journals/corr/abs-2509-23661}. No single model dominates across all inputs, motivating adaptive per-request selection.

\section{INAR-VL}
INAR-VL is an input-aware routing system for edge--cloud VLM inference. In the deployment studied here, the router chooses among four complementary models: two compact edge VLMs that fit within an 8\,GB device budget and two larger cloud VLMs that provide stronger visual and reasoning capability. Given an $(\text{image},\text{question})$ request, INAR-VL constructs a per-request Pareto frontier over a configuration space spanning model architecture, numeric precision, input resolution, and compute placement. The router enumerates valid configurations, scores each on predicted answer quality and estimated cost, and then selects among Pareto-optimal candidates using a calibrated edge-preference rule.

\subsection{System Overview}

\begin{figure*}[!t] 
    \centering
    \includegraphics[width=0.85\textwidth]{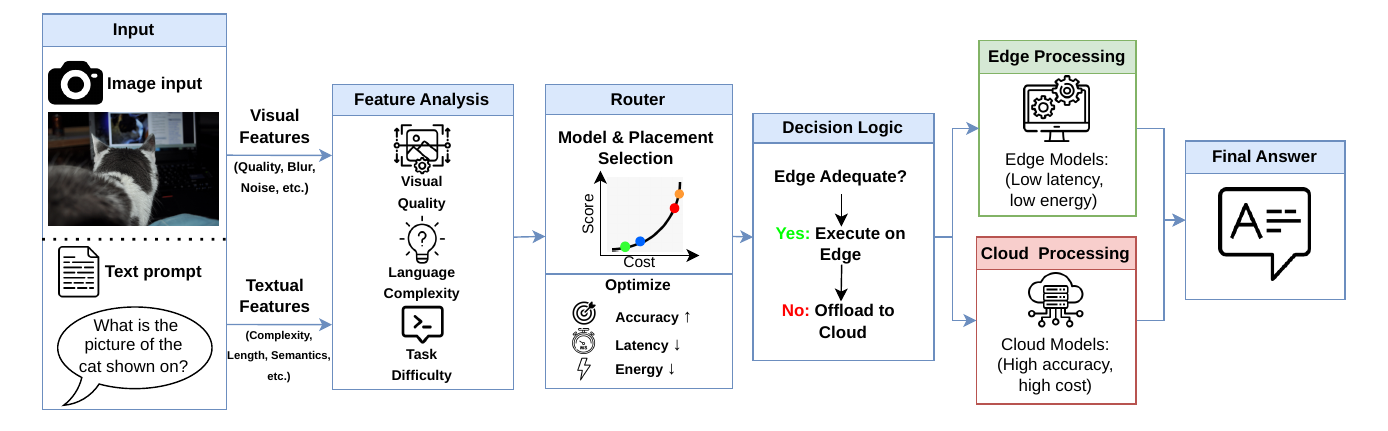}
    \caption{INAR-VL architecture. A multimodal request is routed via a Pareto-based optimizer that selects the model and execution location (edge or cloud) to balance accuracy, latency, and energy.}
    \Description{System architecture of INAR-VL. An image-question request enters the edge node, where lightweight image and text feature extraction runs in parallel and produces a fused feature vector. A Pareto-based router evaluates edge and cloud candidates by predicted quality and estimated cost, selects one execution path, dispatches to the chosen model, and returns the answer.}
    \label{fig:INAR-VL_architecture}
\end{figure*}

\begin{table}[t]
\centering
\scriptsize 
\setlength{\tabcolsep}{3pt} 
\renewcommand{\arraystretch}{1.0} 
\caption{Model pool used by INAR-VL.}
\label{tab:models}
\begin{tabular}{lcccc}
\toprule
\textbf{Model} & \textbf{Params} & \textbf{Deploy} & \textbf{Prec.} & \textbf{Vision Encoder} \\
\midrule
Qwen3-VL-2B & 2B & Edge  & INT8 & Qwen + DeepStack \\
SmolVLM-2B  & 2B & Edge  & INT8 & SigLIP \\
LLaVA-OV-8B & 8B & Cloud & FP16 & RICE-ViT \\
Qwen3-VL-8B & 8B & Cloud & FP16 & Qwen + DeepStack \\
\bottomrule
\end{tabular}
\vspace{-6pt}
\end{table}

Figure~\ref{fig:INAR-VL_architecture} shows the INAR-VL architecture used in our evaluation, and Table~\ref{tab:models} summarizes the model pool used by the router. We follow the pipeline from left to right. (1) \textbf{Input ingestion:} the request is provided as an $(\text{image},\text{question})$ pair. (2) \textbf{Feature extraction:} the image branch computes blur, exposure, artifact, and detail signals. The text branch computes six complexity signals (length, entity/numeric density, question type, vocabulary complexity, reasoning markers, and context-dependency markers). (3) \textbf{Feature analysis and fusion:} both branches are normalized and merged into a 10-dimensional request descriptor $\mathbf{r}$. (4) \textbf{Router:} the router jointly selects model and placement by optimizing predicted accuracy against latency and energy, approximating the Pareto frontier over edge and cloud candidates. (5) \textbf{Decision:} the system prioritizes edge execution when the predicted quality is sufficient, and otherwise offloads to the cloud only if the expected quality gain justifies the additional cost. (6) \textbf{Execution:} the selected candidate is dispatched to either an edge model (INT8) or a cloud model (FP16). (7) \textbf{Response:} the model answer is returned to the client. The system comprises an edge device (NVIDIA RTX 4060, 8\,GB VRAM) and a cloud GPU (NVIDIA RTX PRO 6000, 98\,GB VRAM), and the full routing pipeline adds 10--30\,ms of CPU-side overhead.

Each model carries three normalized strength descriptors in $[0,1]$ (blur robustness, detail/OCR capability, and reasoning ability), which are used by the router's quality predictor to penalize model--input mismatches. These descriptors are fixed offline (not learned online during routing): we initialize them from architecture-informed priors (encoder family, model scale, and known OCR/reasoning characteristics) and verify their relative ordering using single-model pilot measurements on the same calibration protocol used for $a_m$. Values are then frozen for all experiments.

Numerically, ``blur robustness'' denotes the model’s normalized tolerance to blur demand, as reflected in the gap term.
$g_{\text{blur}}=\max(0,\text{need}_{\text{blur}}-\text{strength}_{m,\text{blur}})$. A model with strength $0.7$ incurs no blur-gap penalty until the request's blur need exceeds $0.7$, while a model with strength $0.4$ is penalized earlier on the same input. We use these three axes because they capture the dominant and separable failure modes observed in our model pool: degraded visual perception (blur), fine-grained visual extraction (detail/OCR), and language-side reasoning load.

\section{METHODOLOGY AND CALIBRATION}

\subsection{Request Feature Extraction}

Before invoking any model, INAR-VL extracts a 10-dimensional feature vector $\mathbf{r}\in[0,1]^{10}$, comprising four image-quality axes and six text-complexity axes. We select 10 features as a coverage--efficiency trade-off: this set captures the dominant multimodal variance (visual degradations and language-side reasoning load) while keeping routing overhead within the 10--30\,ms CPU budget. A smaller feature set reduces routing accuracy and under-represents key failure modes (e.g., OCR/detail stress and reasoning-heavy prompts), whereas a larger one increases preprocessing and calibration complexity with diminishing returns, making 10 a practical balance for real-time deployment.
\paragraph{Image quality proxies} 
The image analyzer produces four normalized scores—blur, exposure, JPEG artifacts, and detail—summarized as:
{\setlength{\abovedisplayskip}{2pt}
 \setlength{\belowdisplayskip}{2pt}
\[
s_{\text{img}} = \frac{1}{4}\,(s_{\text{blur}} + s_{\text{exp}} + s_{\text{art}} + s_{\text{detail}})
\]
}
where each $s_{\cdot}\in[0,1]$ and $s_{\text{img}}$ is their mean.

\paragraph{Text complexity} 
The text analyzer computes six normalized axes $C_i\in[0,1]$, combined linearly:
{\setlength{\abovedisplayskip}{2pt}
 \setlength{\belowdisplayskip}{2pt}
\[
c_{\text{text}} = \sum_{i=1}^{6} \beta_i C_i
\]
}
with fixed weights $\beta_i$ from offline calibration.

The joint query complexity is then
{\setlength{\abovedisplayskip}{2pt}
 \setlength{\belowdisplayskip}{2pt}
\[
d = w_{\text{img}}(1-s_{\text{img}}) + w_{\text{txt}}\,c_{\text{text}}
\]
}
where $w_{\text{img}}=w_{\text{txt}}=0.5$ by default. Here, $(1-s_{\text{img}})$ represents image-side complexity and $d\in[0,1]$ denotes overall query complexity; lower $d$ = lower complexity, higher $d$ = higher complexity. In analyses, we report quintiles Q1 (lowest)–Q5 (highest).
\subsection{Pareto Configuration Router}

Each request evaluates configurations $c=(m,p,r,\ell)$ (model, precision, resolution, placement). In the evaluated deployment, precision is fixed by tier, so the active routing choices are model, placement, and edge resolution. The predicted quality of a candidate is
\begin{equation}
\begin{aligned}
\hat q(c,\mathbf r) &= a_m + \delta_{\text{prec}}
+ 0.07\big(\log_2 r-\log_2 r_{\text{cal}}\big) \\
&\quad - \kappa 
\sum_{i \in \{\text{blur},\text{detail},\text{reasoning}\}} g_i^2 
- d(1-a_m) + \Delta_{\text{detail-res}}
\end{aligned}
\end{equation}
where $g_i=\max(0,\text{need}_i-\text{strength}_{m,i})$
, so $\text{strength}_{m,i}$ acts as a capability threshold on axis $i$: penalties appear only when the request's need exceeds the model's normalized capability on that axis.
and $\Delta_{\text{detail-res}}\le 0$ penalizes low-resolution settings on
high-detail images. Cost estimation combines latency, energy, and network terms:
\begin{equation}
\hat c =
B_m f_{\text{prec}}\!\left(\frac{r}{336}\right)^2
+ 0.3\,e(m,p,\ell)
+ n(\ell,bw,r)
\end{equation}

To avoid ambiguity, we define all terms used above:
\begin{itemize}
\item $a_m$: calibrated accuracy of model $m$ measured on a held-out calibration split.
\item $r_{\text{cal}}$: calibration resolution used for $a_m$ (model-specific maximum supported resolution).
\item $\delta_{\text{prec}}$: precision adjustment relative to the calibration precision (inactive in our reported deployment, since precision is fixed by tier).
\item $\kappa$: coefficient for mismatch penalty; we use $\kappa{=}1.0$.
\item $B_m$: base latency factor for model $m$ (proportional to model size in billions of parameters).
\item $f_{\text{prec}}$: precision latency multiplier ($1.0$ for FP16, $0.65$ for INT8).
\item $e(m,p,\ell)$: energy proxy; on edge, $e{=}\text{params}_m\times f_{\text{prec}}$, and on cloud, $e{=}0.1$ (small fixed edge-perspective cost).
\item $n(\ell,bw,r)$: network term; $n{=}0$ for edge and 
$n{=}0.5\linebreak\left(\frac{r}{336}\right)^2\frac{100}{\max(bw,10)}$ for cloud.
\end{itemize}

Parameter calibration follows a one-time offline procedure. We first calibrate $a_m$ on a fixed held-out set (seeded split, same protocol across all models), then fit the remaining coefficients under monotonic constraints (higher resolution should not reduce quality; larger need-strength gaps should not improve quality; higher-complexity queries should hurt weaker models more). The final coefficients ($0.07$, $\kappa{=}1.0$, detail-resolution penalty scale, and cost weights) are then frozen and reused for all datasets and all sweeps.

For all experiments, calibration is performed once per deployment using a held-out set of 1,000 samples per benchmark. The full calibration procedure takes approximately 45 minutes on the edge device and is only required when hardware or model configurations change. This ensures that the routing policy is tailored to the specific deployment environment without incurring runtime overhead during inference.

The router sorts candidates by estimated cost, retains only configurations with strictly increasing predicted quality, and forms a per-request Pareto frontier. It then compares the best edge and best cloud Pareto candidates. Offloading occurs only when the best edge option falls below its calibrated operating point and the best cloud option improves predicted quality by more than $\delta_{\min}=0.03$. This makes the edge preference explicit rather than hiding it in a tie-break rule.

\section{EVALUATION}

We evaluate INAR-VL on three VQA benchmarks and six routing strategies, including three ablations, using the same 2,000-sample subsets. Results report means ± standard deviations and are compared to the oracle.

\subsection{Experimental Setup}

\paragraph{Hardware.} Edge: NVIDIA RTX 4060 (8 GB VRAM). Cloud: NVIDIA RTX PRO 6000 Blackwell (98 GB VRAM), accessed via SSH tunnel. End-to-end cloud inference latency is approximately 2,300 ms, of which about 2,000 ms comes from tunnel and network overhead; the server GPU itself completes 8B FP16 inference in 210--323 ms (see Table~\ref{tab:latency_breakdown} in the Appendix). We report this split to make it clear that, in our setup, the dominant cloud cost is communication rather than GPU execution.

\paragraph{Deployment-transfer scope.} While reported quantitative gains are specific to our edge--cloud benchmark, the INAR-VL framework is hardware-agnostic. Routing depends on request features and calibrated cost/ quality profiles, not on a fixed device type. In our codebase, the same pipeline supports CPU-only edge execution and alternative GPU edge profiles through configuration changes and recalibration of latency/energy priors. 

\paragraph{Deployment Configuration.} Edge models run in INT8 (to fit two models in 8\,GB VRAM) and cloud models in FP16. INAR-VL supports model, precision, resolution, and placement choices, but in our two-tier setup, precision and cloud resolution are fixed, leaving model selection, placement, and edge resolution as routing options. We evaluate six strategies: Edge-Only and Cloud-Only route all requests to one tier; INAR-VL uses 10 features across 4 models; Text-Only and Image-Only use only text or image features; Static routes Qwen-2B to edge and Qwen-8B to cloud.

\paragraph{Energy Measurement} 
Energy is measured via NVIDIA NVML power readings sampled every 50\,ms. Dynamic energy is  
$E = \int \!\max(0, P(t)-P_{\text{idle}})\,dt$,  
where $P(t)$ is instantaneous GPU power and $P_{\text{idle}}$ is average idle power. The $\max(0,\cdot)$ term prevents negative contributions from noise. For cloud execution, GPU energy is measured on the server and returned with the response.

\paragraph{Datasets.}
We evaluate on three visual question answering benchmarks: VQAv2 (val2014, 40.5K samples) \cite{DBLP:journals/ijcv/GoyalKASBP19}, TextVQA \cite{DBLP:conf/cvpr/SinghNSJCBPR19}, and GQA \cite{DBLP:conf/cvpr/HudsonM19}. We use the official releases and sample 2,000 instances per dataset using seeded reservoir sampling.

\paragraph{Metrics.}
Accuracy is measured using the standard VQA soft score $\min(1,\text{matches}/3)$ for VQAv2 and TextVQA, and exact match for GQA. We also report routing split (edge vs. cloud), average latency per request (ms), and average energy per request (J), measured via NVML on both tiers.

\paragraph{Fair-comparison controls.}
All baseline comparisons use the same sampled requests, scoring protocol, and edge/cloud infrastructure. Results reported from prior work are included only as contextual references, as they are obtained under different hardware and software setups and are not strictly comparable.

\paragraph{Strategies compared.}

For context, we compare against an unachievable upper-bound baseline, Oracle (Best-of-4), which selects the per-sample best answer from all four models.

\subsection{Routing Comparison (Main Results)}

Table~\ref{tab:routing} summarizes accuracy across all strategies and datasets, while Table~\ref{tab:latency_energy} reports the corresponding latency, energy, and edge-routing fractions.

\begin{table}[t]
\centering
\footnotesize
\setlength{\tabcolsep}{3pt}
\renewcommand{\arraystretch}{0.9}
\caption{Routing-strategy accuracy ($n{=}2{,}000$ per dataset). MoA-Off reports VQAv2 only.}
\label{tab:routing}
\begin{tabular}{lcccc}
\toprule
Strategy & VQA2 & TextVQA & GQA & Mean \\
\midrule
Edge-Only & 73.7 & 69.0 & 56.8 & 66.5 \\
Cloud-Only & 80.6 & 81.9 & 60.6 & 74.4 \\
\textbf{INAR-VL} & \textbf{79.1} & \textbf{77.4} & \textbf{59.9} & \textbf{72.1} \\
Text-Only & 78.8 & 76.5 & 59.0 & 71.4 \\
Image-Only & 75.1 & 72.4 & 58.0 & 68.6 \\
Static routing & 76.9 & 77.4 & 59.2 & 71.2 \\
MoA-Off & 77.5 & -- & -- & -- \\
\midrule
\emph{Oracle} & \emph{97.4} & \emph{97.0} & \emph{77.5} & \emph{90.6} \\
\bottomrule
\end{tabular}
\end{table}

\textbf{Key findings.}

\begin{figure}[t]
  \vspace{-10pt} 
  \centering
  \includegraphics[width=0.9\columnwidth]{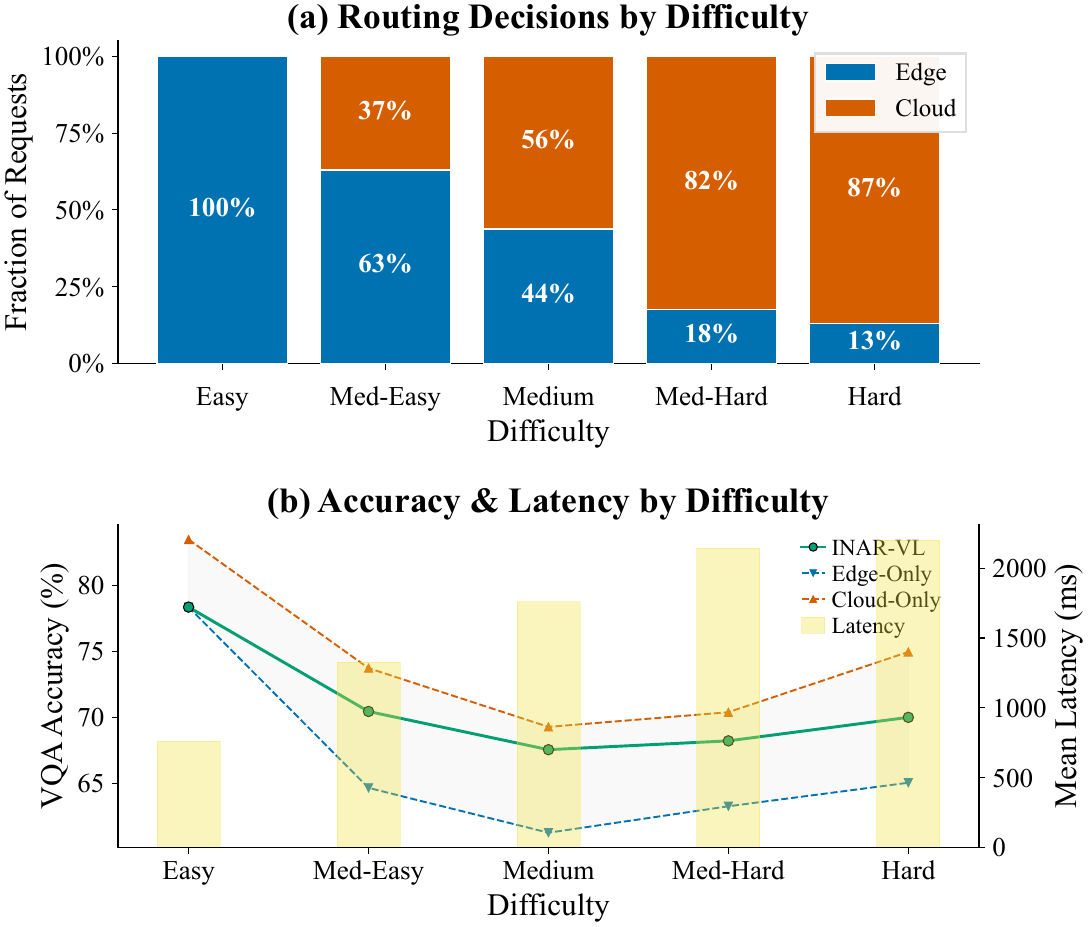}
  \vspace{-6pt}  
  \caption{INAR-VL complexity-aware routing.}
  \Description{Routing distribution and accuracy vs. complexity across edge and cloud execution.}
  \label{fig:complexity_routing}
  \vspace{-10pt} 
\end{figure}

\paragraph{(1) INAR-VL provides the strongest mean accuracy among the evaluated routing variants.}
As shown in Table~\ref{tab:routing}, the INAR-VL router reaches 72.1\% mean accuracy, recovering 71\% of the edge-to-cloud accuracy gap (5.6 of 7.9 pp). Table~\ref{tab:latency_energy} shows that it does so while still serving 36\% of requests on the edge. Figure~\ref{fig:complexity_routing} explains this aggregate result at the per-request level: the router keeps 100\% of Q1 and 63\% of Q2 (lower-complexity) requests on the edge, then shifts a larger share of higher-complexity requests to the cloud, preserving edge efficiency on simpler inputs while narrowing the accuracy gap to cloud-only on more complex ones.

INAR-VL achieves the highest mean accuracy among routing variants (72.1\%), recovering 71\% of the edge-to-cloud accuracy gap (Table~\ref{tab:routing}). As shown in Table~\ref{tab:latency_energy}, it still executes 36\% of requests on the edge. Figure~\ref{fig:complexity_routing} shows that low-complexity queries (Q1–Q2) are routed mostly to the edge, while higher-complexity inputs are offloaded to the cloud, preserving edge efficiency while narrowing the accuracy gap.

\paragraph{(2) Text features dominate routing signal.}
Table~\ref{tab:ablation} shows that the text-only router (71.4\%) matches the full heuristic within 0.7 pp, whereas image-only (68.6\%) barely exceeds always-edge.

\paragraph{(3) Multi-model diversity modestly improves the trade-off surface.}
Restricting the pool to Qwen architectures only (Static routing, 71.2\%) reduces mean accuracy by 0.9\,pp relative to the full four-model INAR-VL (72.1\%) while shifting the edge fraction from 36\% to 31\% in Table~\ref{tab:latency_energy}. This result suggests that architectural diversity improves workload coverage.

\paragraph{(4) The oracle gap remains important context.}
Table~\ref{tab:routing} shows that the oracle (best-of-4) achieves 90.6\% mean accuracy, 18.5\,pp above INAR-VL. This gap exists because the oracle exploits perfect hindsight to select the best model for each sample, while INAR-VL relies on lightweight, pre-inference features and cannot access ground-truth answers or run all models at runtime. As a result, rare or subtle failure modes may go undetected, and the system cannot fully match oracle performance, but aims to recover as much diversity benefit as practical.

Taken together, Tables~\ref{tab:routing}, \ref{tab:ablation}, \ref{tab:latency_energy}, along with Figures~\ref{fig:complexity_routing} and \ref{fig:network}, support the main systems claim of the paper: INAR-VL recovers much of the accuracy gap to cloud-only while reducing latency and energy relative to cloud-only execution, with routing overhead small enough not to erase the benefit.

\begin{table}[t]
\centering
\footnotesize
\setlength{\tabcolsep}{3pt}
\begin{tabular}{lcccc}
\toprule
Ablation & Signal & Acc. & $\Delta$ & Edge\% \\
\midrule
INAR-VL & Img+Txt & 72.1 $\pm$ 1.6 & --- & 36 \\
Text-Only & Text & 71.4 $\pm$ 1.7 & -0.7 & 24 \\
Image-Only & Image & 68.6 $\pm$ 1.7 & -3.5 & 78 \\
Static & Img+Txt & 71.2 $\pm$ 1.6 & -0.9 & 31 \\
Edge Base & None & 66.5 $\pm$ 1.8 & -5.6 & 100 \\
\bottomrule
\end{tabular}
\caption{Ablation of routing signals.}
\label{tab:ablation}
\vspace{-15pt}
\end{table}

\begin{table}[t]
\centering
\footnotesize
\setlength{\tabcolsep}{3pt}
\begin{tabular}{lcccc}
\toprule
Strategy & Lat (ms) & Energy (J) & Acc. & Edge\% \\
\midrule
Edge & 824 $\pm$ 15 & 7.4 $\pm$ 0.6 & 66.5 $\pm$ 1.8 & 100 \\
Cloud & 2408 $\pm$ 125 & 26.0 $\pm$ 1.5 & 74.4 $\pm$ 1.6 & 0 \\
\textbf{INAR-VL} & \textbf{1826 $\pm$ 157} & \textbf{19.2 $\pm$ 1.2} & \textbf{72.1 $\pm$ 1.6} & \textbf{36} \\
Text & 2102 $\pm$ 150 & 21.1 $\pm$ 1.2 & 71.4 $\pm$ 1.7 & 24 \\ 
Image & 1181 $\pm$ 106 & 12.0 $\pm$ 1.1 & 68.6 $\pm$ 1.7 & 78 \\
Static & 1954 $\pm$ 156 & 20.7 $\pm$ 1.2 & 71.2 $\pm$ 1.6 & 31 \\
\bottomrule
\end{tabular}
\caption{Latency, energy, and accuracy. Edge\% indicates edge execution share.}
\label{tab:latency_energy}
\vspace{-10pt}
\end{table}

\subsection{Network Robustness}

We evaluate robustness by sweeping bandwidth from 10\,Mbps (constrained cellular) to 1{,}000\,Mbps (fast LAN) over 6{,}000 samples. For cloud-routed requests, we add an image-transfer overhead of $250\,\text{KB}/\text{bandwidth}$ to latency. The router enforces a bandwidth guard at $b_{\min}=15\,\text{Mbps}$, below which all requests are processed on the edge. Above this threshold, cloud offloading is allowed, with higher costs at lower bandwidth. Accuracy remains stable across bandwidth settings once the guard is satisfied, while latency increases mainly at low bandwidth due to transfer overhead (e.g., $+133$\,ms at 15\,Mbps vs.\ $+7$\,ms at 300\,Mbps).

\begin{figure}[t]
  \centering
  \includegraphics[width=0.7\columnwidth]{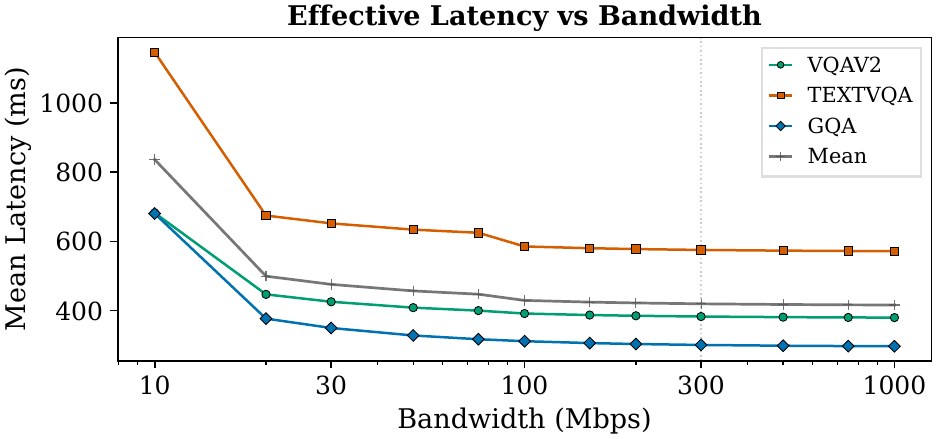}
  \vspace{-4pt}  
  \caption{Impact of bandwidth on latency.}
  \Description{Bandwidth sensitivity of routing latency. Below 15 Mbps, requests remain local; above, offloading reduces latency.}
  \label{fig:network}
\end{figure}

Figure~\ref{fig:network} shows latency improving from 499,ms at 20,Mbps to 416,ms at 1{,}000,Mbps, a 17\% reduction. This indicates that transfer overhead is secondary to inference and fixed communication costs. Below the 15,Mbps guard, the system operates in edge-only mode; once exceeded, latency decreases smoothly with increasing bandwidth.

\section{RELATED WORK}

Edge--cloud collaboration for VLM inference has been explored in systems such as MoA-Off \cite{DBLP:journals/corr/abs-2509-16995}, which dynamically offloads multimodal workloads based on modality complexity, and CE-CoLLM \cite{DBLP:conf/icws/JinW25}, which combines quantization and early-exit techniques to enable collaborative edge--cloud inference. Other systems study task scheduling across heterogeneous deployments. HybridFlow \cite{DBLP:journals/corr/abs-2512-22137} decomposes queries into sub-tasks and routes them to edge or cloud models based on resource availability and expected utility. A related line of work focuses on routing across multiple LLMs to optimize cost--quality trade-offs. Systems such as FrugalGPT \cite{DBLP:journals/tmlr/ChenZ024}, Smoothie \cite{DBLP:conf/nips/GuhaCCKR24}, and RouteLLM \cite{DBLP:journals/corr/abs-2406-18665} learn routers that select between weaker and stronger models based on prompt difficulty. Similarly, Efficient Routing of Inference Requests across LLM Instances in Cloud-Edge Computing \cite{DBLP:journals/corr/abs-2507-15553} formulates routing as a multi-objective optimization problem balancing latency, cost, and response quality, while GreenServ \cite{DBLP:journals/corr/abs-2601-17551} introduces context-aware routing to improve energy efficiency. Table~\ref{tab:related_diff} summarizes key differences. In contrast to prior work that typically optimizes a single objective or modality, INAR-VL jointly routes across heterogeneous VLMs and placements using both image and text signals, explicitly mapping routing decisions to measured latency and energy outcomes.

\begin{table}[H]
\vspace{-8pt}
\centering
\small
\renewcommand{\arraystretch}{0.85}
\setlength{\abovecaptionskip}{3pt}
\setlength{\belowcaptionskip}{0pt}
\begin{tabular}{lccc}
\toprule
Method & Multimodal & Multi-VLM & Cost \\
\midrule
RouteLLM & \xmark & \xmark & \xmark \\
MoA-Off & \cmark & \xmark & \xmark \\
CE-CoLLM & \cmark & \xmark & \xmark \\
HybridFlow & \xmark & \xmark & \cmark \\
GreenServ & \xmark & \xmark & \cmark \\
\textbf{INAR-VL (ours)} & \textbf{\cmark} & \textbf{\cmark} & \textbf{\cmark} \\
\bottomrule
\end{tabular}
\caption{Method-level differentiation.}
\label{tab:related_diff}
\vspace{-15pt}
\end{table}

\section{CONCLUSION}
We presented INAR-VL, a practical edge--cloud routing system for multimodal inference that combines cheap request characterization with an edge-preferring Pareto decision rule. In a real two-tier deployment, the system recovers most of the accuracy advantage of cloud-only execution while improving the deployment trade-off relative to static placements. The additional analyses show why: INAR-VL keeps low-complexity requests local, reserves cloud offloading for complex requests, and maintains nearly the same routing behavior once bandwidth is high enough to make cloud execution feasible. Although our evaluation uses a specific model pool and hardware configuration, the approach generalizes to other settings: new hardware tiers or models can be incorporated by recalibrating the routing thresholds, without changing the core method.

Future work includes extending INAR-VL to more diverse model pools and uncertainty-aware escalation policies that can better exploit the remaining oracle gap.

\begin{acks}
This research was funded in part by the Austrian Science Fund (FWF) through the following projects: \textit{Transprecise Edge Computing (Triton)}, Grant ID: 10.55776/P36870, \textit{Trustworthy and Sustainable Code Offloading (Themis)}, Grant ID: 10.55776/PAT1668223, and by the Vienna Science and Technology Fund (WWTF) through the project \textit{AI-supported Holographic Environmental Water Monitoring (HoloWaterAI)}, Grant ID: ESR24-053.
\end{acks}

\bibliographystyle{ACM-Reference-Format}
\bibliography{Bibliography}

\appendix
\section{Additional Results and Details}

\begin{table}[H]
\centering
\footnotesize
\setlength{\tabcolsep}{6pt}
\renewcommand{\arraystretch}{1.2}
\caption{Per-model GPU latency (ms). Cloud measurements exclude network overhead.}
\label{tab:latency_breakdown}
\begin{tabular}{lccc}
\toprule
\textbf{Model} & \textbf{Deployment} & \textbf{Mean (ms)} & \textbf{P95 (ms)} \\
\midrule
Qwen3-VL-2B & Edge  & 682  & 990  \\
SmolVLM-2B  & Edge  & 1468 & 1846 \\
LLaVA-OV-8B & Cloud & 281  & 371  \\
Qwen3-VL-8B & Cloud & 222  & 282  \\
\bottomrule
\end{tabular}
\end{table}

\begin{table}[H]
\centering
\footnotesize
\setlength{\tabcolsep}{6pt}
\renewcommand{\arraystretch}{1.25}
\caption{Routing strategies evaluated in our experiments.}
\label{tab:strategies}
\begin{tabular}{@{}p{0.27\linewidth} p{0.67\linewidth}@{}}
\toprule
\textbf{Strategy} & \textbf{Description} \\
\midrule
Edge-Only   & All requests are processed on edge devices. \\
Cloud-Only  & All requests are routed to cloud-based models. \\
INAR-VL     & Pareto-based routing using 10 features across 4 models. \\
Text-Only   & INAR-VL variant using only text features ($w_{\text{img}} = 0$). \\
Image-Only  & INAR-VL variant using only image features ($w_{\text{txt}} = 0$). \\
Static      & Fixed routing using Qwen-2B (edge) and Qwen-8B (cloud). \\
\bottomrule
\end{tabular}
\end{table}
\end{document}